\documentclass[11pt,a4paper]{article}
\usepackage[hyperref]{acl2017}
\usepackage{times}
\usepackage{latexsym}
\usepackage{times}
\usepackage{latexsym}
\usepackage{natbib}
\usepackage{graphicx}
\usepackage{multirow}
\usepackage{amsmath,amssymb,amsfonts,amsthm}
\usepackage{url}
\usepackage{xspace}
\usepackage{booktabs}

\newcommand{\ent}{e}
\newcommand{\types}{T}
\newcommand{\alltypes}{\mathcal{T}}
\newcommand{\sentence}{x}
\newcommand{\featurizer}{\varphi}
\newcommand{\weights}{w}
\newcommand{\predictor}{f}
\newcommand{\basepredictor}{f_{\mathrm{single}}}
\newcommand{\figer}{{\sc Figer}\xspace}

\aclfinalcopy 
\def\aclpaperid{3282} 

\title{Fine-Grained Entity Typing with High-Multiplicity Assignments}

\author{Maxim Rabinovich \and Dan Klein\\
  Computer Science Division \\
  University of California, Berkeley \\
  {\tt \{rabinovich,klein\}@cs.berkeley.edu}
}

\date{}

\begin{document}

\maketitle

\begin{abstract} 
As entity type systems become richer and more fine-grained, we expect the number of types assigned to a given entity to increase. However, most fine-grained typing work has focused on datasets that exhibit a low degree of type multiplicity. In this paper, we consider the high-multiplicity regime inherent in data sources such as Wikipedia that have semi-open type systems. We introduce a set-prediction approach to this problem and show that our model outperforms unstructured baselines on a new Wikipedia-based fine-grained typing corpus.
\end{abstract}

\section{Introduction}

Motivated by potential applications to information retrieval, 
coreference resolution, question answering, and other downstream tasks, recent work on entity typing has moved beyond coarse-grained systems towards richer ontologies with much more detailed information, and therefore correspondingly more specific types~\citep{Lin12Figer,Gil14Fine,Yog15Embed}. 

As types become more specific, entities will tend to belong to more types (i.e.\ there will tend to be higher type multiplicity).  However, most data used in previous work exhibits an extremely 
\emph{low} degree of multiplicity. 

In this paper, we focus on the high multiplicity case, which we argue naturally arises
in large-scale knowledge resources. To illustrate this point, we construct a corpus
of entity mentions paired with higher-multiplicity type assignments. Our corpus is based on mentions and categories drawn from Wikipedia, but we generalize and denoise the raw Wikipedia categories to provide more coherent supervision. Table~\ref{tab:examples} gives examples of type assignments from our dataset.

As type multiplicity grows, it is natural to consider type prediction as an inherently set-valued problem and ask questions about how such sets might be modeled.  To this end, we develop a structured 
prediction approach in which the sets of assigned types are predicted as first-class 
objects, including a preliminary consideration of how to efficiently search over them. The resulting model captures type correlations and ultimately outperforms a
strong unstructured baseline.

\begin{table*}
\centering
\begin{tabular}{cc}
David Foster Wallace & \texttt{novelist suicide sportswriter writer alumnus ...} \\
Albert Einstein & \texttt{physicist agnostic emigrant people pacifist ...} \\
NATO & \texttt{organization treaty document organisation alliance ...} \\
Federal Reserve & \texttt{agency authorities banks institution organization ...} \\
Industrial Revolution & \texttt{concept history evolution revolution past ...} \\
Black Death & \texttt{concept epidemic pandemic disaster ...} 
\end{tabular}
\caption{With types from a large corpus like Wikipedia, large type set assignments become common.\label{tab:examples}}
\end{table*}

\begin{table*}
\centering
\begin{tabular}{ccc}
\toprule
Entity & Raw Type & Projected Type \\ \midrule
\multirow{5}{*}{David Foster Wallace} & \texttt{Short\_story\_writers} & \texttt{writer} \\
 & \texttt{Amherst\_alumni} & \texttt{alumnus} \\
 & \texttt{Illinois\_State\_faculty} & \texttt{faculty} \\
 & \texttt{People\_from\_New\_York} & \texttt{people} \\
 & \texttt{Essayists} & \texttt{essayist} \\
\end{tabular}
\caption{Example of an entity and its types, before and after projection. The projection operation collapses related types that would be very difficult to learn in their original, highly specific forms.}
\end{table*}

\begin{figure}[t]
\includegraphics[width=\linewidth]{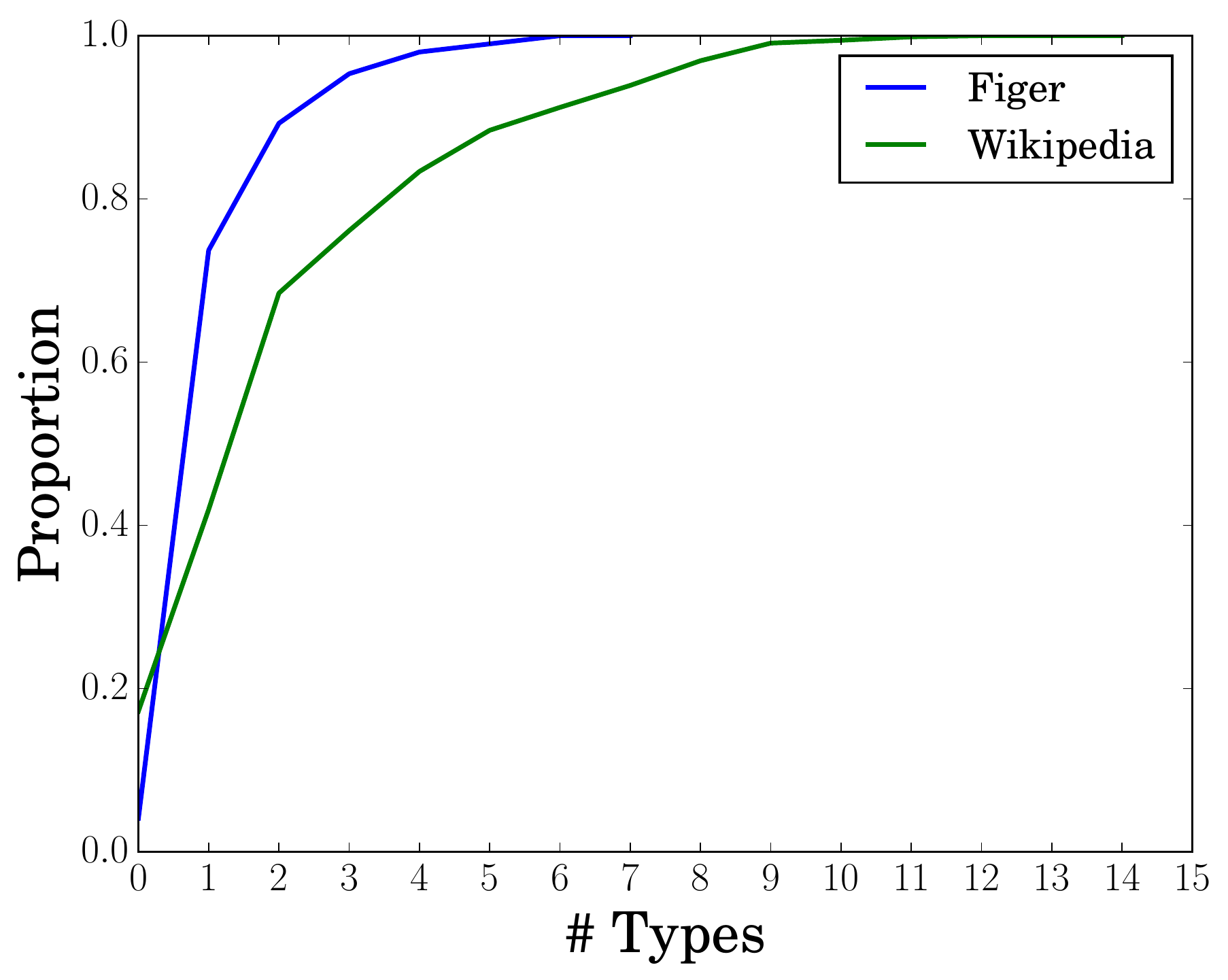}
\caption{Comparison of type set size CDFs for the our Wikipedia corpus and the prior \figer corpus~\citep{Lin12Figer}. The figure illustrates that our corpus exhibits much greater type assignment multiplicity.}
\end{figure}

\paragraph{Related work}

The fine-grained entity typing problem was first investigated in detail 
by~\citet{Lin12Figer}. Subsequently,~\citet{Gil14Fine} introduced a larger evaluation 
corpus for this task and introduced methods for training predictors based on multiclass 
classification. Both used the Freebase typing system, coarsened to approximately $100$ types, and subsequent work has mostly followed this lead~\citep{Yag16Corpus,Yog15Embed},
although types based on WordNet have recently also been investigated~\citep{Cor15Finet}.

Most prior work has focused on unstructured predictors using some form of multiclass 
logistic regression~\citep{Lin12Figer,Gil14Fine,Shi16Attentive,Yag16Corpus,Yog15Embed}.
Some of these approaches implicitly incorporate structure during decoding by enforcing
hierarchy constraints~\citep{Gil14Fine}, while neural approaches can encode correlations
in a soft manner via shared hidden layers~\citep{Shi16Attentive,Yag16Corpus}.

Our work differs from these lines of work in two respects: its use of a corpus exhibiting high type multiplicity with types derived from a semi-open inventory and its use of a fully structured model and decoding procedure, one that can in principle be integrated with neural models if desired. Previously, most results focused on the low-multiplicity Freebase-based \figer corpus. The only work we are aware of that uses a type system similar to ours used a rule-based system and evaluated on their own newswire- and Twitter-based evaluation corpora~\citep{Cor15Finet}.

\section{Model}

Our structured prediction framework is based on modeling type assignments as sets.
Each entity $\ent$ is assigned a set of types $\types^{\ast}$ 
drawn from the overall set of types $\alltypes$. Our goal is thus to predict, given an 
input sentence-entity pair, the set of types associated with that entity.

We take the commonly-used linear model approach to this structured prediction problem.
Given a featurizer $\featurizer$ that takes an input sentence $\sentence$ and entity 
$\ent$, we seek to learn a weight vector $\weights$ such that
\begin{align}\label{eq:f-general}
\predictor\left(\sentence,~\ent\right) & = 
\mathrm{argmax}_{\types}~ \weights^{\top}\featurizer\left(\sentence,~\ent,~\types\right) 
\end{align}
predicts $\types$ correctly with high accuracy.

Our approach stands in contrast to prior work, which deployed several techniques, of
similar efficacy, to port single-type learning and inference strategies to the multi-type setting~\citep{Gil14Fine}. Provided type interactions can be neglected, equation~\eqref{eq:f-general} can be simplified to
\begin{align*}
\basepredictor\left(\sentence,~\ent\right) & = 
\left\lbrace t \in \alltypes \colon ~
\weights^{\top}\featurizer\left(\sentence,~\ent,~t\right) \geq r \right\rbrace .
\end{align*}
This simplification corresponds to expanding each multi-type example triple $\left(\sentence,~\ent,~\types^{\ast}\right)$
into a set of single-type example triples $\left\lbrace \left(\sentence,~\ent,~t^{\ast}\right)_{t^{\ast} \in \types^{\ast}}\right\rbrace$. 
Learning can then be done using any technique for multiclass logistic regression, and 
inference can be carried out by specifying a threshold $r$ and predicting all types that score above that threshold: In prior work, a simple $r = 0$ threshold was used~\citep{Lin12Figer}.

In this paper, we focus on the more general specification~\eqref{eq:f-general}, though
in Section~\ref{subsec:learn-infer}, we explain a simplification that can be used to speed up inference if desired.

\subsection{Features}\label{subsec:features}

Modeling type assignments as sets in principle opens the door to non-decomposable
set features (a simple instance of which would be set size). For reasons of 
tractability, we assume our features factor along type pairs:
\begin{equation}\label{eq:features}
\featurizer\left(x,~e,~T\right) = \sum_{t \in T} \featurizer\left(x,~e,~t\right) + \sum_{t,~t' \in T} \featurizer\left(t,~t'\right)    
\end{equation}
Note that in addition to enforcing factorization over type pairs, the 
specification~\eqref{eq:features} requires that any features linking the type assignment to the observed entity mention depend only on a single type at a time. We investigated
non-decomposable features, but found they did not lead to improved performance.

We use entity mention features very similar to those in previous work:
\begin{enumerate}

\item {\bf Context unigrams and bigrams.} Indicators on all uni- and bigrams 
within a certain window of the entity mention.

\item {\bf Dependency parse features.} Indicators on the lexical parent
of the entity mention head, as well as the corresponding dependency type. Separately, indicators on the lexical children of the entity mention head and their 
dependency types.

\item {\bf Entity head and non-head tokens.} Indicators on the
syntactic head of the entity mention and on its non-head tokens.

\item {\bf Word shape features.} Indicators on the shape of each token in the 
entity mention.

\end{enumerate}

We combine these features with type-based features to obtain the features our model actually uses:
\begin{enumerate}

\item {\bf Conjunction features.} These are simple conjunctions of mention features with
indicators on type membership in the predicted set. Using only these features results in an 
unstructured model.

\item {\bf Type pair features.} These are indicators on pairs of types appearing in the
predicted set.

\item {\bf Graph-based features.} As we discuss in Section~\ref{sec:corpus},
the type system in our corpus comes with a graph structure. We add indicators on certain patterns occurring within the set--e.g. a parent-child type pair, sibling type pairs, and so on, abstracting away the specific types.

\end{enumerate}

\subsection{Learning and Inference}\label{subsec:learn-infer}

\begin{figure}
\centering
\includegraphics[clip, scale=0.4]{{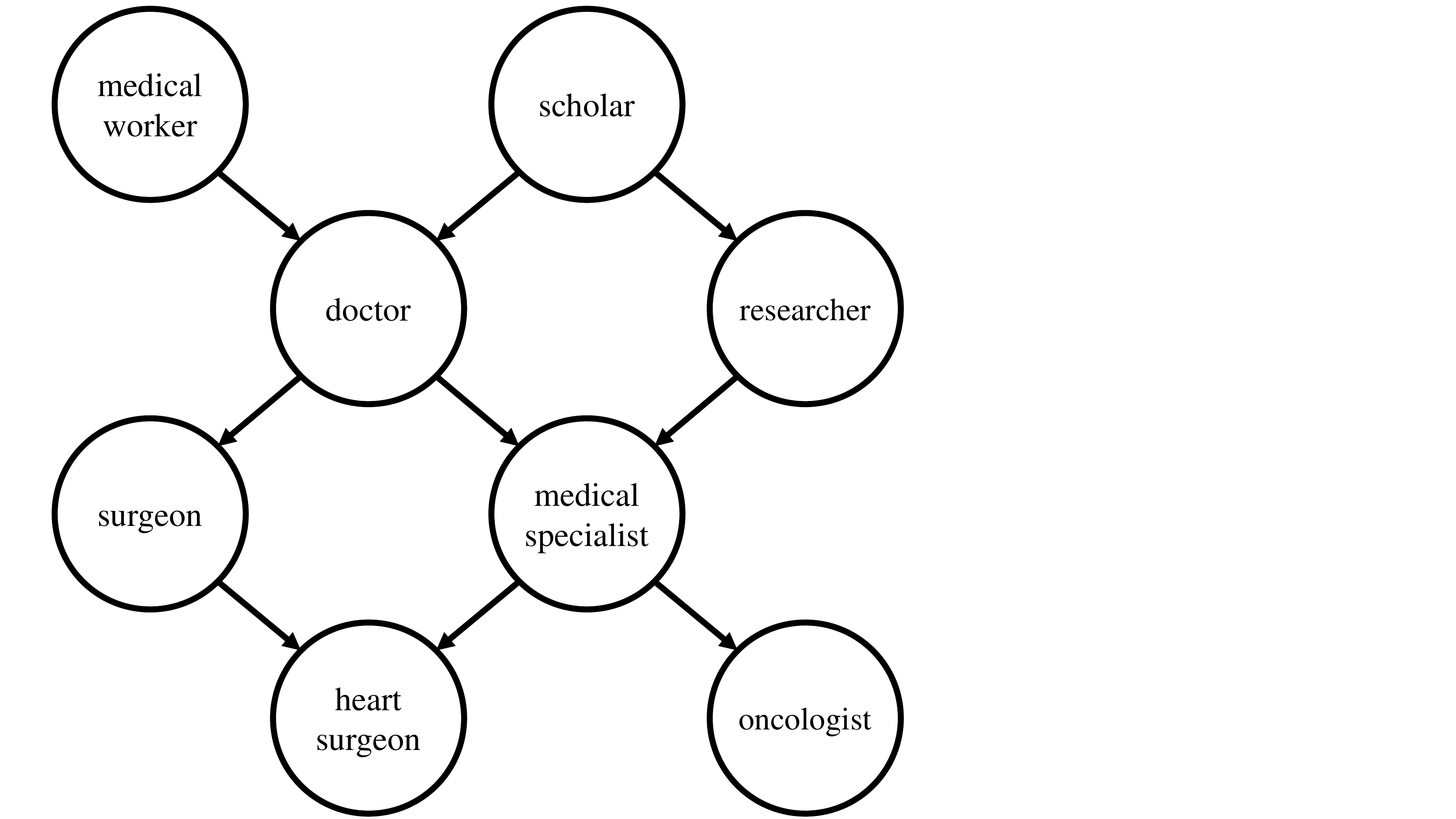}}
\caption{Fragment of the graph underlying our type system.\label{fig:graph-fragment}}
\end{figure}

We train our system using structured max-margin~\citep{Tso05StructMarg}. 
Optimization is performed via AdaGrad on the primal~\citep{Kum15MaxMarg}. 
We use set-F1 as our loss function.

Inference, for both prediction and loss-augmented decoding, poses a greater challenge,
as solving the maximization problem~\eqref{eq:f-general} exactly requires iterating over all subsets of the type system. 

Fortunately, we find a simple greedy algorithm is effective. Our decoder begins by
choosing the type that scores highest individually, taking only single-type features 
into account. It then proceeds by iteratively adding new types into the set until doing
so would decrease the score. 

At the cost of restricting the permissible type sets slightly, we can speed up the greedy procedure further. Specifically, we can require that the predicted type set $T$ be \emph{connected} in some constraint graph over the types---either the co-occurrence graph, the complete graph, or the graph underlying the type system. If we denote by $\mathcal{C}$ the set of all such connected sets, the corresponding predictor would be
\begin{equation*}    
\predictor_{\mathrm{conn}}\left(\sentence,~\ent\right) = 
\mathrm{argmax}_{\types \in \mathcal{C}}~ \weights^{\top}\featurizer\left(\sentence,~\ent,~\types\right) 
\end{equation*}
The greedy decoding procedure for this predictor is faster because at each step, it need only consider adding types that are adjacent to some type that has already been included.

\section{Corpus}\label{sec:corpus}

\begin{table}[t]
\begin{tabular}{ll|ccc}
Level & Features & P & R & F1 \\ \hline
Entity & Unstructured & 50.0 & {\bf 67.2} & 52.9 \\ 
& ~~ + Pairs & 53.3 & 64.1 & 54.3 \\ 
& ~~ + Graph & {\bf 53.9} & 63.9 & {\bf 54.5} \\ \hline
Sentence & Unstructured & 42.6 & {\bf 58.9} & 44.4 \\ 
& ~~ + Pairs & 46.5 & 54.1 & {\bf 45.6} \\ 
& ~~ + Graph & {\bf 47.0} & 53.6 & {\bf 45.6} \\
\end{tabular}
\caption{Results on our corpus. All quantities are macro-averaged.\label{tab:results}}
\end{table}

Our corpus construction methodology involves three key stages: mention identification,
type system construction, and type assignment.\footnote{Our corpus will be released at \url{http://people.eecs.berkeley.edu/~rabinovich/}.} We explain each of these in turn. 

\paragraph{Mention identification.} We follow prior work on entity linking~\citep{Dur14Joint} and take all mentions that occur as anchor text. We filter the resulting collection of mentions down to those that pass a heuristic filter that removes mentions of common nouns, as well as spurious sentences representing Wikipedia formatting.

\paragraph{Type system construction.} Prior work on fine-grained entity typing has
derived its type system from Freebase~\citep{Lin12Figer,Gil14Fine}. 
The resulting ontologies thus 
inherit the coverage and specificity limitations of Freebase, somewhat exacerbated by 
manual coarsening.

Motivated by efforts to inject broader coverage, more complex knowledge resources into NLP systems, we instead derive our types from the Wikipedia category and WordNet graphs, in a manner similar to that of~\citet{Pon07Deriving}. 

Our base type set consists of all Wikipedia categories. By following back-pointers in
articles for categories, we derive a base underlying directed graph. To eliminate noise, we filter down to all categories whose syntactic heads can be found in WordNet and keep directed edges only when the head of the parent is a WordNet ancestor of the head of the child. We conclude by projecting each type down to its syntactic head. 

\paragraph{Type assignment.} The type set for an entity is obtained by taking its
Wikipedia category assignments, augmenting these with their ancestors in the category
graph above, and then projecting these down to their syntactic heads.

\section{Experiments}

\begin{figure}[t]
\centering
\includegraphics[width=\linewidth]{{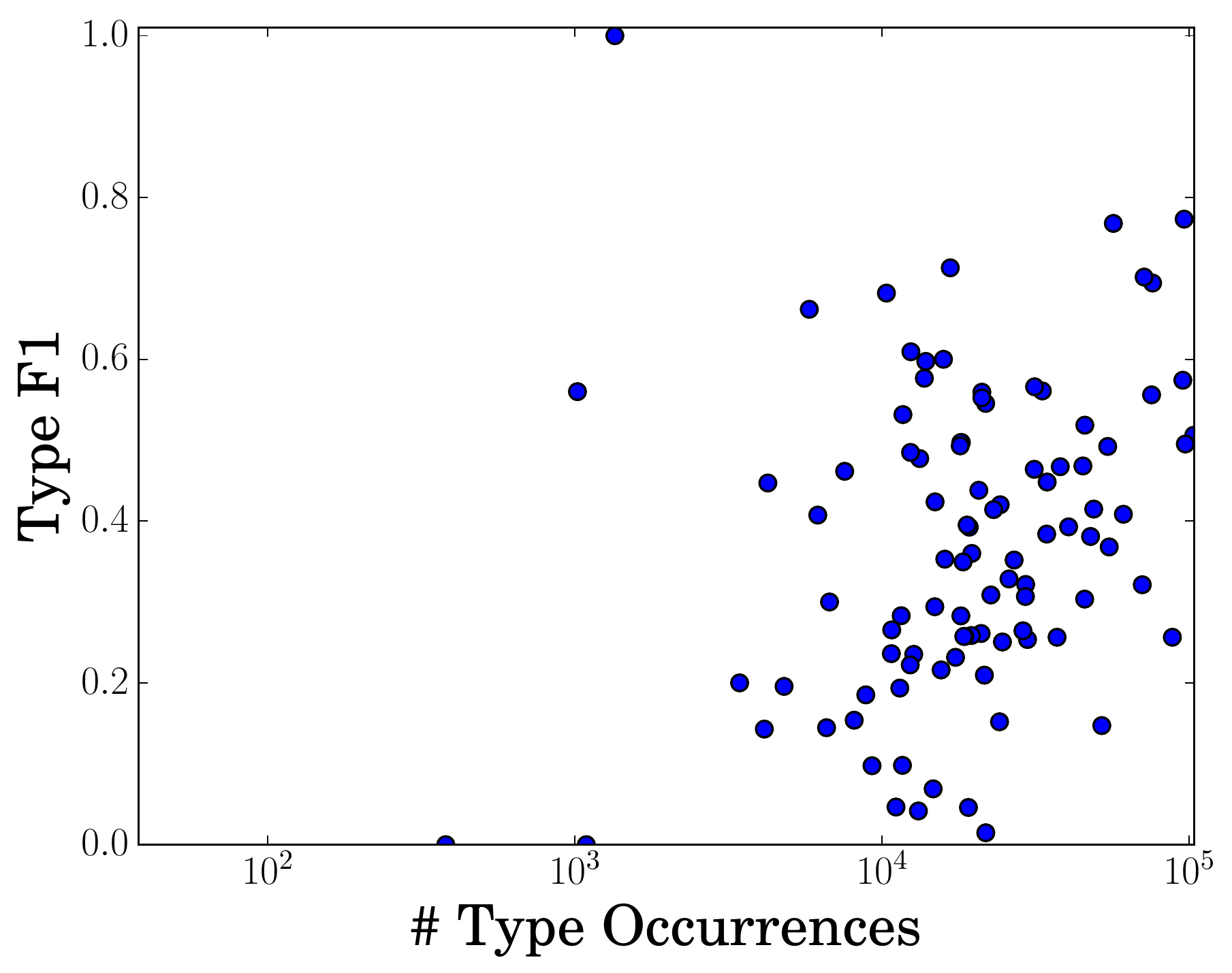}}
\caption{Per-type F1 scores plotted by type frequency in the training corpus.\label{fig:results-by-type}}
\end{figure}

We evaluate our method on the dataset described in Section~\ref{sec:corpus}. For these experiments, we restrict to the
$100$ most frequent types and downsample to 750K mentions. We use a baseline that closely replicates the \figer system~\citep{Lin12Figer}. Within our framework, this can be thought of as a model that sets all type pair features in~\eqref{eq:features} to zero.


Table~\ref{tab:results} summarizes our results. Starting with the baseline, we incrementally add the type pair, graph-based, and set size features discussed in~\ref{subsec:features}. Adding type pair features results in an appreciable  performance gain, while the graph features bring little benefit---potentially because pairwise correlations suffice to summarize the set structure when the number of types is moderately low.

A concern when studying multiclass problems with large numbers of classes, whether predicting sets or individual labels, is that performance on instances associated with common classes will dominate the performance metric. Figure~\ref{fig:results-by-type} shows micro-averaged F1 for the binary prediction task associated with predicting the presence or absence of each type, demonstrating that our performance is strong even for many rare types.

\section{Conclusion}
%
%
%
%

We have highlighted the issue of multiplicity in fine-grained entity typing.
Whereas most prior work has focused on corpora with low multiplicity assignments, 
we denoised the Wikipedia type system to construct a realistic 
corpus with high multiplicity type assignments. Using this corpus as a testbed, we showed that an approach based on structured prediction of sets can outperform 
unstructured baselines when type assignments have high multiplicity. Our approach may 
therefore be preferable in such contexts. 


\newpage

\bibliography{paper}
\bibliographystyle{acl_natbib}

\end{document}